\documentclass[pdflatex,sn-vancouver-num]{sn-jnl}
\usepackage{graphicx}%
\usepackage{multirow}%
\usepackage{amsmath,amssymb,amsfonts}%
\usepackage{amsthm}%
\usepackage{mathrsfs}%
\usepackage[title]{appendix}%
\usepackage{xcolor}%
\usepackage{textcomp}%
\usepackage{manyfoot}%
\usepackage{booktabs}%
\usepackage{algorithm}%
\usepackage{algorithmicx}%
\usepackage{algpseudocode}%
\usepackage{listings}%
\usepackage{tikz}
\usetikzlibrary{positioning}
\usepackage{pgf-pie}
\usepackage[utf8]{inputenc}
\usepackage{tabularx}
\usepackage{setspace}
\usepackage[section]{placeins}

\newcolumntype{L}[1]{>{\raggedright\arraybackslash}p{#1}}
\theoremstyle{thmstyleone}%

\theoremstyle{thmstyletwo}%
\theoremstyle{thmstylethree}%
\begin{document}
\title[An LLM Pipeline for EHR Documentation Inconsistencies]{Toward Automated Detection of Documentation Inconsistencies in Electronic Health Records}

\author[1]{\fnm{Jian} \sur{Lu}, BS}\email{jian.lu@duke.edu}

\author[1]{\fnm{Panyu} \sur{Chen}, MS}\email{panyu.chen@duke.edu}

\author[1]{\fnm{Miriam} \sur{Treggiari}, MD, PhD}\email{miriam.treggiari@duke.edu}

\author[1]{\fnm{Robert} \sur{Blessing}, MD}\email{robert.blessing@duke.edu}

\author[1]{\fnm{Danyang} \sur{Zhuo}, PhD}\email{danyang.zhuo@duke.edu}

\author[2]{\fnm{Chunhua} \sur{Weng}, PhD}\email{cw2384@cumc.columbia.edu}

\author[3]{\fnm{William W.} \sur{Stead}, MD}\email{bill.stead@vumc.org}

\author*[1]{\fnm{Anru R.} \sur{Zhang}, PhD}\email{anru.stat@gmail.com}
\affil[1]{\orgname{Duke University}, \orgaddress{\street{2424 Erwin Road}, \city{Durham}, \postcode{27705}, \state{North Carolina}, \country{USA}}}
\affil[2]{\orgname{Columbia University Medical Center}, \orgaddress{\street{622 West 168 Street}, \city{New York}, \postcode{10032}, \state{NY}, \country{USA}}}
\affil[3]{\orgname{Vanderbilt University Medical Center}, \orgaddress{\street{2525 West End Avenue}, \city{Nashville}, \postcode{37203}, \state{TN}, \country{USA}}}
\abstract{
\textbf{Objective:} To characterize the kinds of internal documentation inconsistencies a general-domain large language model (LLM) can surface from real-world discharge summaries, and to identify recurring failure modes that limit reliability at scale.

\textbf{Materials and Methods:} We applied a two-stage LLM pipeline---open-ended candidate identification (Gemini 2.5 Pro) followed by context-grounded verification (Gemini 2.5 Flash)---to 3{,}000 MIMIC-IV-Note discharge summaries drawn as three contiguous 1{,}000-note batches from distinct positions in the corpus. Two clinicians independently reviewed a domain-stratified sample of 268 retained candidates; a third adjudicated disagreements.

\textbf{Results:} Stage~1 proposed 11{,}333 candidates; 3{,}461 were retained after Stage~2 verification---machine-generated candidates---with at least one in 69.7\% of admissions. Medication (43.2\%) and diagnosis or clinical-finding (25.3\%) candidates predominated. In the reviewed sample, reviewers agreed on 91.0\% of candidates, and 253 of 268 (94.4\%) were judged valid after adjudication. Representative examples from all seven domains carried direct implications for clinical reasoning or patient safety. Review also revealed recurring failure modes involving temporal reasoning, evolving diagnoses, and outpatient-prescribing conventions.

\textbf{Discussion:} Detection is highly context-dependent: judging a flagged pair requires locating each statement and assessing whether the conflict reflects a true contradiction or missing context. We propose a graded ontology spanning strict contradiction, ambiguity, and gap, with a schema characterizing each flagged case by type, section, scope, domain, attribute, and inconsistency axis.

\textbf{Conclusion:} This formative study establishes a methodological foundation and conceptual framework to guide subsequent validated, large-scale EHR-inconsistency analysis.
}
\maketitle

\section{Background and Significance}\label{sec:background}
Electronic Health Records (EHRs) are fundamental for modern healthcare, supporting a wide range of downstream tasks, such as clinical care management,\cite{komorowski2018artificial} clinical outcome prediction,\cite{harutyunyan2019multitask, gentimis2017predicting} medical named-entity recognition,\cite{guo2024detection} and patient phenotyping.\cite{yang2024enhancing} However, the reliability of information within EHR datasets is not infallible, with prior studies documenting pervasive issues such as incomplete documentation, internal inconsistencies, and clinically meaningful errors across care settings.\cite{weiskopf2013methods, Hogan2015, balogh2015improving, graber2017impact} These documentation errors can propagate through clinical workflows and contribute to diagnostic errors, which affect an estimated 5\% of outpatient encounters.\cite{singh2014frequency} Error of documentation is a widely recognized safety risk and can lead to inappropriate or delayed care.\cite{jointcommission2018copy} In addition, common documentation workflows such as copy-paste and templating can propagate outdated or incorrect information across notes, creating internal contradictions and, in documented cases, contributing to missed diagnoses and patient harm.\cite{jointcommission2014copypaste,weir2003direct,colicchio2019unintended} Medication discrepancies are another example: inconsistencies in medication histories and discharge documentation can persist across transitions of care and contribute to adverse drug events that are not reliably caught by routine electronic checks.\cite{cornish2005unintended,koppel2005cpoe} Empirical evidence of such inconsistencies in clinical practice has so far come primarily from manual review: in a large survey of ambulatory note readers, Bell et al.~\cite{bell2020patient} found that approximately one in five patients reported perceiving an error in their notes, with more than 40\% of reported errors judged to be serious or very serious. While compelling, such analyses depend on patient self-report and manual chart inspection, and do not scale to systematic characterization of EHR documentation quality across large cohorts.

Large language models (LLMs) have recently emerged as a powerful general-purpose technology for natural-language understanding, with strong performance demonstrated across a wide range of domains including law,\cite{guha2023legalbench} finance,\cite{wu2023bloomberggpt} and medicine.\cite{thirunavukarasu2023llmmedicine,clusmann2023future} In clinical applications specifically, LLMs have been used for medical question answering at expert-level performance,\cite{singhal2023medpalm,singhal2025medpalm2} clinical information extraction from unstructured notes,\cite{agrawal2022clinicalIE,yang2022gatortron} and clinical text summarization.\cite{vanveen2024clinicalsumm} For automated medical-error detection, the MEDIQA-CORR shared task~\cite{abacha2024mediqacorr,benabacha2025medec} has spurred a range of LLM-based approaches, including retrieval-augmented prompt programs,\cite{wang2024mediqacorr} error-category-aware ensembles,\cite{gundabathula2024promptmind} and chain-of-thought prompting with in-context hints.\cite{gema2024edinburgh} These approaches have proven effective in benchmark settings where the underlying errors are manually injected into short, curated clinical snippets. They differ substantially from the inconsistency-detection task we consider, which operates on real-world EHR datasets such as MIMIC-IV,\cite{johnson2020mimic} where discharge summaries span thousands of words and integrate information across diagnoses, medications, procedures, laboratory results, and care plans. Recent LLMs are increasingly able to handle long-context inputs, yet it remains an open question whether they can reliably surface inconsistencies in real-world clinical documentation, and what failure modes emerge when they cannot.

In this work, we address these questions through a formative study of a two-stage LLM detection pipeline, open-ended candidate identification followed by context-grounded verification, applied to full-length discharge summaries from a large MIMIC-IV cohort. The pipeline surfaces candidate inconsistencies across demographics, allergies, procedures, diagnoses, laboratory, medications, and care-planning domains, many of which could carry serious downstream implications for clinical reasoning and patient safety. At the same time, we observe recurring failure modes that arise from the complexity of real-world EHR data: inconsistency detection is highly context-dependent, requiring temporal reasoning, clinical-knowledge nuance, and awareness of documentation conventions that general-domain LLMs may not reliably capture. Drawing on these observations, we develop a graded ontology of clinical inconsistency---spanning strict contradiction, ambiguity, and gap---and a schema for characterizing each flagged case. Together, the empirical findings and this ontology-and-schema framework provide a foundation for systematic, validated EHR-inconsistency analysis and a basis for the curated corpus build that the next phase of this work requires.

\section{Methods}
\label{sec:methods}
\subsection{Data Source}
We used MIMIC-IV as the development cohort for all analyses reported in this paper. We focused exclusively on discharge summaries, which serve as integrative documents summarizing demographics, allergies, procedures, diagnoses, laboratory, and medications across a hospital encounter. The primary dataset is the Medical Information Mart for Intensive Care IV (MIMIC-IV),\cite{johnson2020mimic} together with its associated clinical notes module, MIMIC-IV-Note,\cite{johnson2023mimicnote} from which discharge summaries are drawn. MIMIC-IV is a large, publicly available, de-identified EHR dataset spanning hospital and intensive care unit admissions at a single tertiary academic medical center, and comprises longitudinal structured clinical data for over 330{,}000 hospital admissions; MIMIC-IV-Note provides the corresponding free-text clinical notes, including discharge summaries authored by clinicians at the end of each admission. From the full MIMIC-IV-Note discharge summary corpus, we sampled 3{,}000 discharge summaries, drawn as three contiguous 1{,}000-note batches from distinct positions in the corpus. Because the corpus is ordered by patient identifier, a contiguous batch can contain several admissions of the same patient. We treat this as a formative sample appropriate for qualitative characterization of pipeline behavior; quantitative analysis is reserved for the curated-corpus development (Section~\ref{sec:futurework}).

\subsection{Ethics and Data Access}
\label{subsec:ethics}
Analysis of the publicly available MIMIC-IV dataset,\cite{johnson2020mimic} which contains de-identified clinical records, was conducted in accordance with its data use agreement and did not require additional institutional review board (IRB) approval; study personnel completed the PhysioNet credentialed-access training and signed the MIMIC data use agreement. The study did not involve direct interaction with human subjects.

\subsection{Iterative Pipeline Development Guided by Qualitative Analysis}
\begin{figure}[!htbp]
\centering
\includegraphics[width=\linewidth]{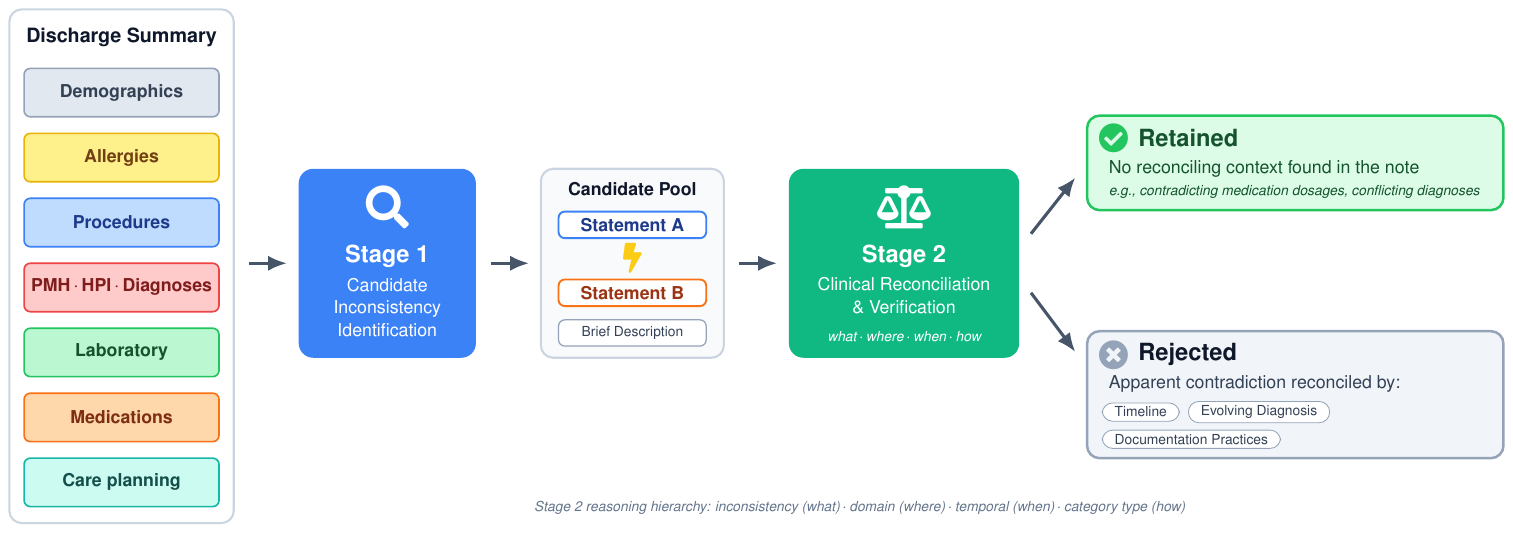}
\caption{Overview of the two-stage inconsistency detection pipeline. Stage~1 scans a full discharge summary and returns a pool of candidate inconsistencies, each a pair of potentially conflicting statements; Stage~2 verifies each candidate against the surrounding clinical context, retaining candidates that the note does not reconcile and excluding apparent ones reconcilable through timing, evolving diagnoses, or documentation conventions.}
\label{fig:pipeline_overview}
\end{figure}

We developed the pipeline architecture iteratively, beginning with a minimal single-prompt design and adding a verification stage in response to observed limitations. The detection pipeline is summarized in Figure~\ref{fig:pipeline_overview} and serves as the version applied to all results reported in this paper. This subsection traces the iterative development of its two stages: open-ended candidate identification and contextual verification.

\paragraph{Initial design: open-ended detection.}
The initial version of the pipeline consisted of a single LLM call that took a full discharge summary as input and returned a list of candidate inconsistencies (this prompt is retained as \emph{Stage~1} in the final pipeline). It operates in an exploratory mode: the model has no specific claim to evaluate and must scan the entire document for pairs of statements that may disagree. The prompt asks the model to consider inconsistencies across a broad set of clinical categories---medications (presence/absence, dose, frequency, indication); diagnoses and major conditions; laboratory results and imaging with diagnostic implications; clinical course and outcomes; procedures, devices, laterality, and code status; and allergies or contraindications. To reduce noise, it explicitly excludes (i)~differences clearly explained by timing or expected clinical evolution, (ii)~minor numeric or wording differences with the same clinical intent (e.g., ``26 days'' vs.\ ``4 weeks''), (iii)~approximate durations or rounding artifacts,(iv)~apparent conflicts attributable to de-identification placeholders rather than to the clinical content, (v)~redundant phrasing, and (vi)~apparent inconsistencies without impact on clinical interpretation or management. The output is a candidate list in which each entry contains the two conflicting statements and a brief description of why the apparent inconsistency may matter clinically. By design, this stage is optimized for recall: it produces a candidate pool that downstream verification will prune.

\paragraph{Adding verification: inverting the search direction.}
This single-stage configuration surfaced a substantial number of false-positive contradictions: candidate pairs that appeared to disagree in isolation but were reconcilable when read against the surrounding clinical context. Reconciliations frequently involved temporal evolution (laboratory values that legitimately change between time points), evolving diagnostic assessments, or standard documentation conventions (e.g., a drug discontinued during the stay differs between the admission and discharge medication lists). These failures reflect a structural mismatch between what the prompt asks the model to do---divergent search across the whole document---and what verification requires---convergent reasoning against a specific candidate. We therefore added a second stage structurally inverse to the first. \emph{Stage~2} takes a specific candidate pair as input and prompts the model to reason about it step by step \cite{wei2022cot}---identifying what each statement claims, locating it within the discharge summary, and assessing whether timing or broader clinical context could reconcile the two---before returning a binary verdict on whether the apparent inconsistency survives a strict, document-grounded reading. A pair is retained only if the discharge summary provides \emph{no} plausible timeline or clinical explanation that reconciles the two, an asymmetric burden of proof that biases the stage toward conservative, high-precision classification. Where Stage~1 starts without a target and searches a document for candidates, Stage~2 starts with a candidate and searches the same document for reconciling context. This inverse framing substantially reduced false-positive retention relative to the single-stage configuration. Residual cases that pass Stage~2 yet remain reconcilable on closer inspection are characterized in Section~\ref{sec:failure_modes}.

\paragraph{Compute allocation.}
To maximize computational efficiency, we assign different models to the two detection stages based on the required level of reasoning. The more capable (and more expensive) model is reserved for open-ended candidate generation, which requires hypothesis generation and cross-sectional reasoning, while a faster, more cost-efficient model handles context-grounded verification. We used Gemini 2.5 Pro for candidate identification (Stage~1) and Gemini 2.5 Flash for context-grounded verification (Stage~2)\cite{google2025gemini25}. Mean input/output token counts per call were 3{,}565/3{,}618 for Stage~1 and 3{,}892/1{,}263 for Stage~2, with output counts including thinking tokens. The estimated total model cost for processing 3{,}000 discharge summaries through Vertex AI batch prediction, including normalization, was US\$91 (approximately \$0.03 per note).

\subsection{Expert Chart Review Protocol}
\label{sec:chart_review}
To identify clinically useful inconsistencies and characterize the model's strengths and limitations, we conducted a structured expert review of a subset of candidates retained after Stage~2. The review examined which candidates warranted clinical attention and where the model's interpretation fell short. This section describes the review protocol and procedures; Section~\ref{sec:review_results} presents the results.
\paragraph{Sampling strategy.}
\label{subsec:sampling}
From the 3{,}461 candidates retained after Stage~2, we randomly sampled cases for expert review, stratified by clinical domain (twelve labels, grouped into the seven domains of Table~\ref{tab:domain_taxonomy} for reporting). Let $N_d$ denote the number of retained candidates in domain $d$, and $n_d$ the number sampled from it:
\[
n_d = \min\bigl(50,\; \max(10,\; \operatorname{round}(0.1\,N_d))\bigr),
\]
giving 268 candidates in total (7.7\% of those retained). Sampling was performed at the candidate level rather than the admission level, to ensure coverage across clinical domains.
\paragraph{Review task.} For each candidate, reviewers examined the two flagged statements and the model’s description, then judged whether the pair represented a documentation inconsistency (Yes, No, or Unsure) based on their clinical judgment. The excerpts came from the same de-identified text used by the pipeline. Reviewers could also provide free-text comments.

\paragraph{Reviewers and adjudication.} Two clinicians---an anesthesiologist with ten years of experience in anesthesiology, and a critical care nurse practitioner with more than thirty years of clinical experience---labeled all 268 candidates independently, without access to each other's labels. Candidates on which their labels differed were referred to a senior physician and biomedical informatician, whose label was final.

\section{Pipeline Output and Formative Case Analysis}
\label{sec:results}

\subsection[Pipeline Output]{Pipeline Output}
\label{sec:pipeline_output}
Across 3{,}000 discharge summaries, Stage~1 proposed 11{,}333 candidate pairs and Stage~2 retained 3{,}461 (30.5\%), covering 2{,}092 admissions (69.7\%). The median number retained per admission was 1 (IQR 0--2; maximum 6), or 1 (IQR 1--2) among admissions with any candidate. These are pipeline-output counts, not prevalence estimates; expert review covered a sample only (Section~\ref{sec:review_results}).

\begin{figure}[!htbp]
\centering
\includegraphics[width=\linewidth]{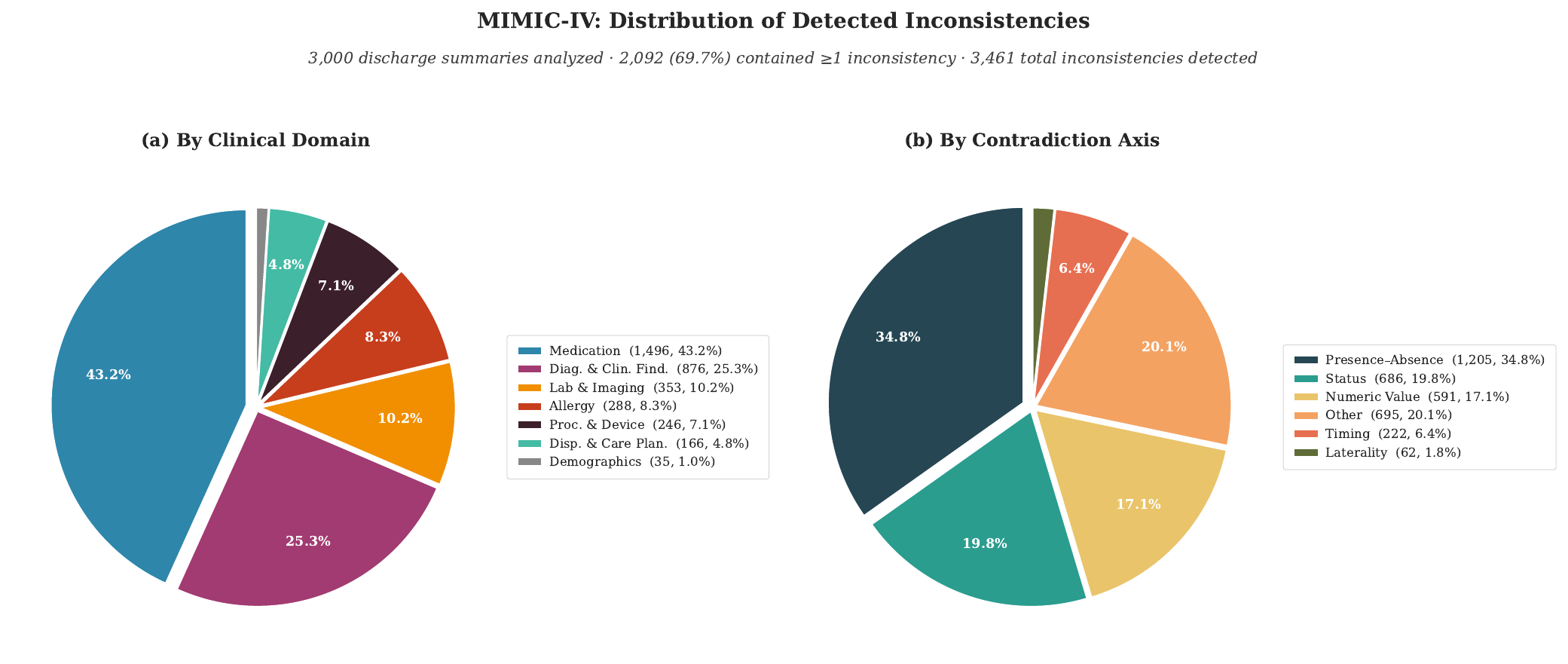}
\caption{Distribution of the 3{,}461 candidates retained after Stage~2 across 3{,}000 MIMIC-IV discharge summaries, by (a)~clinical domain and (b)~contradiction axis. Counts describe pipeline output, not adjudicated inconsistencies.}
\label{fig:mimic_pie}
\end{figure}

Medication and diagnosis or clinical-finding candidates predominated, and presence--absence was the most common contradiction axis (Figure~\ref{fig:mimic_pie}). Section mapping resolved 6{,}428 of 6{,}922 statements (92.9\%); the leading locations were discharge medications (24.0\% of mapped statements), hospital course (18.4\%), and discharge instructions (11.0\%). Among 3{,}060 candidates with both statements mapped, 41.8\% crossed encounter phases, 36.2\% crossed sections within a phase, and 22.1\% occurred within one section.

Although Stage~1 excluded apparent conflicts attributable to de-identification placeholders, we screened all retained candidates by string matching for the placeholder \texttt{\_\_\_}; 637 (18.4\%) contained it in at least one flagged statement. We then screened these candidates with Gemini 2.5 Flash, providing the two statements and the model's explanation to assess whether the apparent inconsistency depended on redacted content. The automated screen classified 54 candidates (1.6\% of all retained candidates) as depending on redacted content, suggesting possible de-identification artifacts.

\subsection[Expert Chart Review]{Expert Chart Review}
\label{sec:review_results}
\paragraph{Agreement and outcome.} The two primary reviewers assigned the same validity label to 244 of the 268 candidates (91.0\%). Their disagreements concerned uncertainty rather than rejection: of the 24 disagreements, 23 paired a Yes with an Unsure, 1 paired a Yes with a No. Cohen's $\kappa$ was low (0.06), a known behavior of $\kappa$ under extreme prevalence imbalance. After adjudication, 253 candidates (94.4\%) were judged valid and 15 (5.6\%) invalid. All 15 invalid judgments came from adjudication of disputed candidates; no candidate was rejected by both primary reviewers.

\paragraph{Interpretation.} These results should be read narrowly. The sample was stratified by domain rather than drawn in proportion to the pipeline's output, so 94.4\% is not an estimate of validity across all 3{,}461 candidates, and the review says nothing about inconsistencies the pipeline missed. The disputed candidates also show that validity is not always settled by the text: the same candidate could appear valid to one clinician and uncertain to another, a pattern we return to in Section~\ref{sec:ontology}. The review was conducted internally, by members of the study team, as part of this formative study. Its goal was to identify failure modes and representative inconsistencies to guide a more rigorous quantitative analysis, not to assess the pipeline's accuracy.

\subsection{Clinically Meaningful Candidate Inconsistencies}
In this section, we present representative examples that illustrate the types of documentation issues the pipeline can surface. We also discuss candidate pairs that, upon clinical review, appear reconcilable and therefore highlight recurring failure modes of the approach (Section~\ref{sec:failure_modes}). The representative cases are organized by clinical domain and follow the typical ordering of content within a discharge summary. Together, they span a spectrum of potential clinical significance, ranging from documentation discrepancies with limited downstream impact to inconsistencies that may directly affect clinical decision and patient safety. Table~\ref{tab:valid_examples} summarizes these cases.
\begin{table}[!htbp]
\centering
\caption{Representative pipeline-detected documentation inconsistency by clinical domain. Cases are ordered by their typical position in a discharge summary.}
\label{tab:valid_examples}
\footnotesize
\setlength{\tabcolsep}{4pt}
\renewcommand{\arraystretch}{1.2}
\begin{tabularx}{\linewidth}{l X X}
\toprule
\textbf{Domain} & \textbf{Inconsistency} & \textbf{Clinical commentary} \\
\midrule
Demographics & HPI refers to patient as ``He''; demographics and
remainder of note consistently use female sex and references. & Likely a
copy-paste error; could cause confusion for future chart readers. \\
\addlinespace

Allergy & Allergy list documents spironolactone; discharge medication
list includes spironolactone 100\,mg PO daily. & Prescribing a medication
to which the patient has a listed allergy is a documented safety risk;
discrepancy requires clarification before dispensing. \\
\addlinespace

Procedure \& device & Discharge summary contains a detailed
``DISCHARGE INSTRUCTIONS FOLLOWING TRANSMETATARSAL AMPUTATION'' section;
the procedure list and hospital course state no amputation was performed
this admission. & Post-amputation instructions for a procedure that did
not occur could prompt inappropriate wound care, weight-bearing
restrictions, and follow-up at the receiving facility. \\
\addlinespace

Diagnosis \& clinical finding & HPI describes a stable patient who
``denies nausea, vomiting, fevers, chills, or any other complaints'';
brief hospital course states the patient was hypotensive to the 80s
systolic on admission requiring fluid resuscitation. & The HPI omits a
critical hemodynamic finding that bears on differential diagnosis
(e.g., sepsis, hemorrhage) and on potential hospital disposition. \\
\addlinespace

Laboratory \& imaging & Brief hospital course reports ``urine culture
grew Klebsiella and Pseudomonas''; pertinent results show
\emph{Klebsiella pneumoniae} and \emph{Proteus mirabilis}. & Misnaming
the causative organism can mislead future antimicrobial selection even
when the in-stay regimen happened to cover both. \\
\addlinespace

Medication & Discharge medication list states ``Amiodarone 200\,mg PO
DAILY''; the prescription detail in the same Discharge Medications block
reads ``RX amiodarone 200\,mg \dots\ by mouth twice a day.'' & The same
medication entry contains two different dosing frequencies (once daily
vs.\ twice daily), creating a risk of dosing error with implications
for amiodarone toxicity\\
\addlinespace

Disposition \& care planning & Code status documented in one section as
``DNR''; another section records that the patient wants extraordinary
measures. & The two statements describe opposite;
in-stay decision-making requires immediate clarification. \\
\bottomrule
\end{tabularx}
\end{table} 
\subsubsection*{Demographics}
Demographic inconsistencies are relatively mild in consequence but notable when they appear, typically reflecting copy-paste errors rather than clinical disagreement. They do not bear directly on clinical management but illustrate how templated note generation can propagate identity-level errors into the chart.

\subsubsection*{Allergy}
Allergy-related inconsistencies arose from copy-paste carry-forward of legacy entries, intersections between drug-class labels and individual agents, or entries that record adverse drug reactions rather than true allergies. All three can leave a documented allergy alongside an active same-class prescription; reconciliation depends on a clinician or pharmacy noticing the link across sections.

\subsubsection*{Procedure \& device}
Procedure- and device-related inconsistencies include a clinically distinctive subset of cases in which the contradiction concerns \emph{laterality}---the side of the body on which a procedure was performed, a device was placed, or a finding was reported. Related cases involve discrepant documentation of implanted or indwelling devices (e.g., whether a central line, drain, or implant is in place at the time of discharge). Errors of this kind, even when reconcilable on careful chart review, are the documentation form most exposed to misdirecting downstream readers who encounter only a fragment of the chart.

\subsubsection*{Diagnosis \& clinical finding}
Diagnostic and clinical-finding inconsistencies can carry direct implications for clinical reasoning when categorical labels disagree across sections of the note. Such disagreements typically appear when no temporal or methodological account in the document reconciles the competing classifications.

\subsubsection*{Laboratory \& imaging}
Inconsistencies between objective laboratory or imaging findings and the clinical narrative were a recurring detection pattern, often arising when a narrative summary lags or contradicts the underlying measurements. Numeric discrepancies across sections, even when small in absolute terms, can plausibly alter how a downstream reader interprets severity.

\subsubsection*{Medications}
Medication-related inconsistencies were a common kind of detection in the MIMIC-IV cohort, frequently arising when an update to one section of a discharge summary was not propagated to others. Even small dosing errors of this kind can introduce confusion at medication reconciliation, particularly for agents with narrow therapeutic windows.

\subsubsection*{Disposition \& care planning}
Disposition- and care-planning-related inconsistencies appear at the interface between the in-stay record and the instructions handed to the patient or receiving facility, frequently involving code status, functional or weight-bearing status, or the discharge destination itself. Because these statements directly govern post-discharge management, even narrow disagreements between sections can translate into unsafe handoffs or actively conflicting orders at the next site of care.
 
Taken together, these cases illustrate the variety of candidate documentation inconsistencies the pipeline surfaces across clinical domains and the range of clinical implications they could plausibly carry. They should be interpreted as formative examples rather than as adjudicated prevalence estimates.

\subsection{Recurring Failure Modes}
\label{sec:failure_modes}
Not every flagged pair, however, represents a genuine documentation inconsistency. On review with clinical experts, a recurring subset of flagged pairs proved reconcilable with clinical context, reflecting systematic patterns in how the language model interprets discharge documentation. We describe three such failure modes below, followed by a fourth category that counts as a failure only under the pipeline's binary definition of inconsistency; Table~\ref{tab:apparent_inconsistencies} summarizes the cases.

\begin{table}[!htbp]
\centering
\caption{Representative pipeline-flagged statement pairs that do not meet the original strict-contradiction criterion, organized by failure-mode category.}
\label{tab:apparent_inconsistencies}
\footnotesize
\setlength{\tabcolsep}{4pt}
\renewcommand{\arraystretch}{1.25}
\begin{tabularx}{\linewidth}{L{0.18\linewidth} L{0.20\linewidth} X X}
\toprule
\textbf{Domain} & \textbf{Failure mode} & \textbf{Apparent inconsistency} & \textbf{Clinical rationale} \\
\midrule
Allergy & Documentation clarity & Allergy list states no known allergies; hospital course documents narcotic sensitivity causing confusion. & The statements do not strictly contradict each other, but the sensitivity documented in the hospital course may indicate a gap in the allergy/adverse-reaction record.\\
\addlinespace
Allergy & Clinical nuance & Simvastatin listed as allergy; rosuvastatin prescribed at discharge. & Statin cross-reactivity is limited; switching within the class is standard practice when the prior reaction is intolerance rather than true allergy. \\
\midrule
Diagnosis \& clinical finding & Clinical nuance & Physical exam documents 2+ DP pulses bilaterally; arterial studies show monophasic waveforms. & Palpable pulses may be maintained by collateral circulation or vessel wall stiffness despite proximal stenosis. \\
\midrule

Lab \& imaging & Temporal progression & MRI reports preserved systolic function; echocardiogram shows EF $\sim$20\%. & Two studies at different time points; likely clinical deterioration between imaging studies. \\
\midrule
Medication & Temporal progression & Hospital course plans a 10-day course of doxycycline; discharge prescription dispenses only 9 tablets (4.5 days at BID dosing). & Patient received $\sim$5.5 days of doxycycline inpatient; the dispensed supply covers the remaining outpatient days of the 10-day total course. \\
\addlinespace
Medication & Documentation convention & Colonoscopy report states polyps not removed because patient was on Plavix; hospital course states Plavix was recently held. & Standard practice is to hold Plavix 5--7 days before a procedure; the reason for deferral is the same regardless of phrasing. \\
\midrule
Disposition \& care planning & Documentation convention & Patient documented as DNR/DNI with comfort measures; discharge instructions advise returning to ER for worsening symptoms. & Standardized discharge instruction templates may not reflect individualized goals of care; not a true internal contradiction. \\
\bottomrule
\end{tabularx}
\end{table}
\subsubsection*{Temporal progression}
A substantial fraction of apparent inconsistencies arise from statements that appear contradictory in isolation but reflect findings or events recorded at different time points during the same admission. A closely related sub-pattern involves reasoning about medication-course durations: when an inpatient and outpatient supply together complete a stated course, the pipeline sometimes flags the dispensed quantity as inconsistent with the planned duration. Both forms of temporal-progression failure---reasoning across event time points and reasoning across medication-course windows---point to a shared refinement target: explicit extraction of a clinical-event timeline (imaging dates, laboratory draw times, prescribed course start/end, inpatient administration history) that downstream comparisons can be evaluated against, rather than treating the discharge summary as a flat block of text.
\subsubsection*{Clinical nuance}
A second category of apparent inconsistencies arises from cases where domain-specific knowledge resolves an apparent conflict — within-class substitution of medications when the prior reaction reflects intolerance rather than true allergy, or palpable peripheral pulses coexisting with abnormal arterial-flow studies in patients with non-compressible vessels. Unlike the other failure modes, where pipeline behavior is the addressable root cause, clinical-nuance cases depend on knowledge external to the discharge summary; closing this gap is less a matter of improving the language model and more a matter of giving the pipeline access to clinical-knowledge resources. Cases of this type require pipeline access to clinical-knowledge resources beyond the discharge summary itself.
\subsubsection*{Documentation conventions}
Standardized discharge instructions and templated phrasing can produce text that appears to conflict with the clinical narrative even when the underlying clinical decision is consistent---a procedural deferral attributable to a medication held according to standard practice, or boilerplate return-to-ER instructions accompanying documented goals of comfort care. Such cases reflect institutional templating rather than internal contradiction, and would be flagged less often by a pipeline aware of common documentation conventions.

\subsubsection*{Actionable documentation gaps}
A final category is not a strict contradiction. These pairs suggest that documentation may be incomplete---a sensitivity documented in the hospital course but never entered on the allergy list, for example. Because the pipeline equated inconsistency with strict contradiction, returning a binary verdict on whether two statements conflict, such pairs count as failures: the statements do not conflict, and labeling them a contradiction mischaracterizes them. Yet they point to a real documentation problem, one of omission rather than conflict. We return to this category in Section~\ref{sec:ontology}, where it motivates treating documentation gaps as a class of their own.

\section{Discussion}
\label{sec:lessons}

The case analysis above suggests that automated inconsistency detection requires more than a binary label of ``consistent'' or ``inconsistent.'' Some flagged pairs are strict contradictions; others are ambiguous, clinically concerning, or reconcilable once timing, clinical knowledge, or documentation conventions are considered. We therefore separate two conceptual tasks. First, we define a graded ontology that specifies what kinds of cases should count as clinically relevant inconsistency-like phenomena. Second, we define a schema for characterizing each flagged case in a structured and reproducible way. The ontology addresses \emph{what type of case this is}; the schema addresses \emph{where it occurs and what it concerns}.

\subsection{A Graded Ontology of Clinical Inconsistency}
\label{sec:ontology}
In the logic of formal consistency,\cite{dowden} a set of statements is \emph{inconsistent} when they cannot all be true together; for a pair, statements~A and~B are inconsistent when they cannot both hold. A \emph{contradiction} is the strong case, in which exactly one of the two must be true and the other false.

Applying this definition to clinical documentation, however, is rarely so clean. A key finding of our formative study is that inconsistency in the clinical domain is highly context-dependent (Section~\ref{sec:failure_modes}). Two laboratory values that differ are not contradictory if they were drawn at different time points and the patient's state evolved between them; an apparent conflict between a documented allergy and a prescribed drug may not be an inconsistency once within-class pharmacology is considered. In each case both statements can be true---the appearance of inconsistency arises from reading them without the timing, clinical knowledge, or documentation conventions a clinician supplies automatically. Still other cases occupy a genuine gray zone: an allergy list recording no known allergies alongside a documented narcotic sensitivity is not a strict contradiction, yet neither is it clearly free of a documentation problem.

These observations motivate a working notion of \emph{clinical inconsistency} that is broader than the strict logical definition. We take it to encompass the class of statement pairs worth flagging in a discharge summary---both contradictions in the strict logical sense and lighter cases that, while not strict contradictions, still warrant clinical attention. Within this class, the cases fall along a gradient of strength. At the strict end, a \emph{strict contradiction} is a pair that cannot both be true under any correct reading of the encounter, so that one statement is necessarily wrong---a procedure recorded on the left side versus the right is the canonical case. Beyond it, the formative study surfaced lighter cases that are not strict contradictions and carry less impact on downstream care, but are still worth flagging. One is \emph{ambiguity}: a pair that is not contradictory but whose documentation is underdetermined---it admits more than one reading, and a reader cannot reliably tell which was intended; the defect is one of clarity rather than of truth, but it can still mislead. These categories form a graded ontology of clinical inconsistency. A third kind of case is not a matter of degree. A \emph{gap} arises when every statement is readable but an item that the documentation itself entails is absent---an abnormal result that is never assessed, or a new opioid prescription with no bowel regimen. Whereas an ambiguity leaves a reader unable to tell what an entry says, a gap leaves the reader knowing exactly what it says and missing what it requires; its repair is a new entry elsewhere in the note rather than a clarification of the existing one. The actionable documentation gaps described in Section~\ref{sec:failure_modes} are cases of this kind. The pipeline equated inconsistency with strict contradiction, and under that binary view these cases were failures: their statements do not conflict. Table~\ref{tab:ontology} summarizes the three classes.

\begin{table}[!htbp]
\centering
\caption{Ontology of clinical inconsistency.}
\label{tab:ontology}
\footnotesize
\setlength{\tabcolsep}{4pt}
\renewcommand{\arraystretch}{1.25}
\begin{tabularx}{\linewidth}{L{2.5cm} X}
\toprule
\textbf{Class} & \textbf{Definition} \\
\midrule
Strict contradiction & Two statements cannot both be true under any correct reading of the encounter, so at least one is necessarily wrong. \\
Ambiguity & The documentation admits more than one reading, and a reader cannot determine which was intended. \\
Gap & The documented statements are interpretable, but an item required by the documentation itself is absent from the note. \\
\bottomrule
\end{tabularx}
\end{table}

\subsection{A Schema for Characterizing a Clinical Inconsistency}
\begin{figure}[!htbp]
\centering
\includegraphics[width=\linewidth]{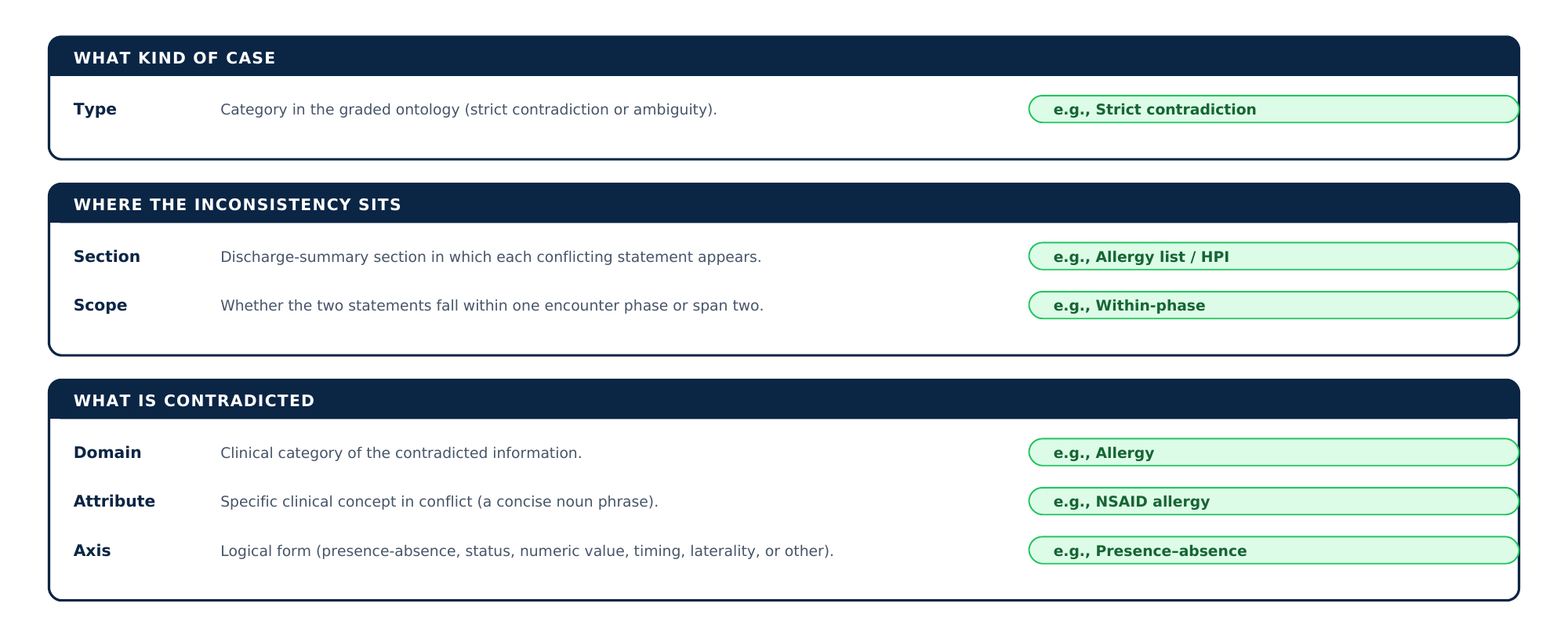}
\caption{Normalization schema for documentation inconsistencies.}
\label{fig:normalization_schema}

\end{figure}

Whereas the ontology defines the type of inconsistency-like phenomenon, the schema defines how each flagged case is represented for systematic review. This distinction is essential because the pipeline surfaces heterogeneous cases that differ not only in clinical domain, but also in document location, temporal scope, and logical structure. The schema summarized in Figure~\ref{fig:normalization_schema} serves two purposes. First, it provides the specificity required for systematic detection, classification, and adjudication of candidate inconsistencies. Second, it creates a structured basis for designing interventions, because different error patterns require different remedies: laterality conflicts, medication-reconciliation discrepancies, temporal-progression artifacts, and templated-instruction conflicts are unlikely to be addressed by the same mechanism. We therefore treat the schema as a working vocabulary to be tested and refined during corpus labeling.

The first dimension is the inconsistency \emph{type}: the case's category in the ontology of Section~\ref{sec:ontology}---strict contradiction, ambiguity, or gap. This field labels each instance at the level of the ontology before the remaining dimensions locate and characterize it within the document, and serves as the link between the ontology and the per-case description that follows.

\begin{figure}[!htbp]
\centering
\includegraphics[width=\linewidth]{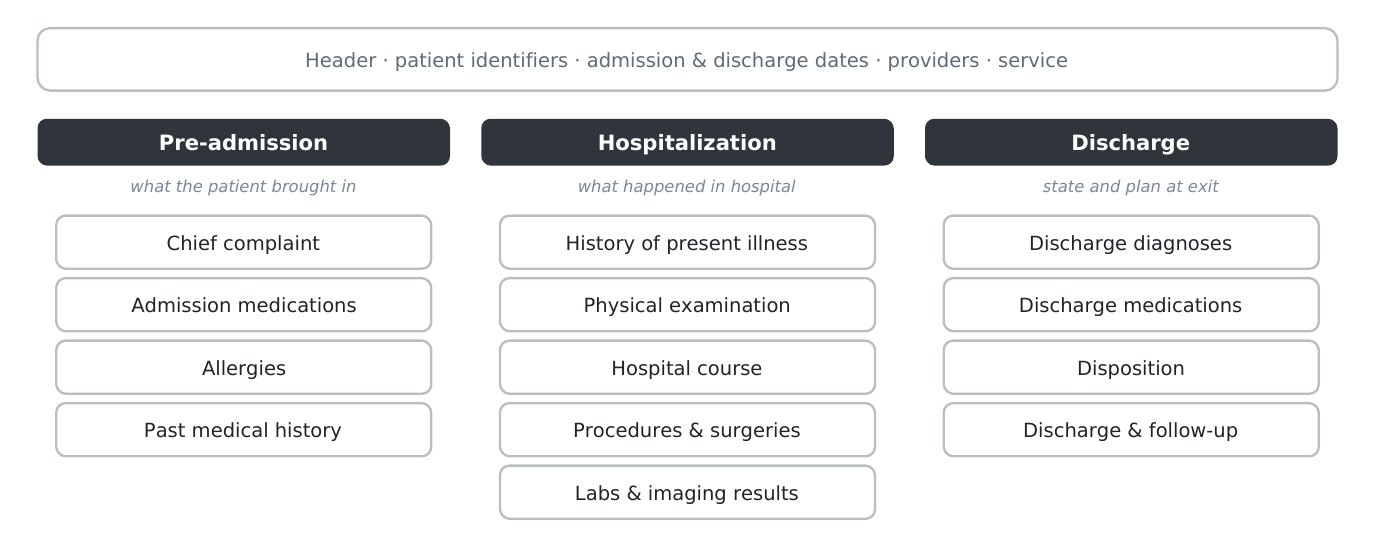}
\caption{Structure of a discharge summary. Beneath an identifying header, sections are organized into three encounter phases: the patient's state before admission, events during hospitalization, and the discharge state and plan. \emph{Admission medications} refer to medications the patient was taking immediately before admission. Medications administered during the hospitalization are considered part of the hospital course. \emph{Discharge medications} refer to medications prescribed or continued after discharge. Dashes indicate sections that are not typically present in that phase rather than missing data. This schematic is anchored to MIMIC-IV-Note discharge summaries but reflects a broadly recognizable discharge-summary structure.}
\label{fig:discharge_anatomy}

\end{figure}

The next two dimensions describe \emph{where} the inconsistency sits in the document, using two levels of structure (Figure~\ref{fig:discharge_anatomy}). A \emph{section} is a headed block of the discharge summary, such as Allergies or Discharge Medications. Sections belong to one of three encounter \emph{phases}: \emph{pre-admission}, the patient's state before this admission (e.g., Chief Complaint, Allergies, Past Medical History); \emph{hospitalization}, what happened during the stay (e.g., Hospital Course, laboratory results); and \emph{discharge}, the patient's state and plan at discharge (e.g., Discharge Medications, Disposition). This arrangement is stable across discharge summaries and broadly similar across institutions. The \emph{section} field records the section in which each conflicting statement appears, and the \emph{scope} field records whether the two statements fall within a single phase or span two. The distinction matters: when statements fall in different phases, an apparent disagreement can often be explained by the patient's legitimate change over time rather than by a documentation error. Such cross-phase pairs warrant caution and should not be treated as inconsistencies without further evidence---a difference between admission and discharge medications is frequently a normal consequence of the hospitalization. The exception is content tied to a \emph{time-invariant attribute}, most clearly an allergy: a patient recorded as allergic to a drug and then treated with it may represent a genuine inconsistency requiring clinical reconcilation.

The remaining dimensions describe \emph{what} is contradicted. The \emph{domain} is the clinical category of the contradicted information---medication, allergy, laboratory and imaging, diagnosis, procedure, and so on (Table~\ref{tab:domain_taxonomy})---and places the inconsistency in clinical space, which is closely tied to its potential for harm and to the clinical workflow involved. Complementary to the domain, and orthogonal to it, is the \emph{contradiction axis}: the logical form the disagreement takes---presence versus absence, conflicting status, conflicting numeric value, conflicting timing, or conflicting laterality (Table~\ref{tab:axis_taxonomy})---independent of the clinical subject matter. Together the two let an inconsistency be named by both clinical area and logical type---a \emph{medication timing} conflict, an \emph{allergy presence--absence} conflict---so that recurring patterns can be aggregated along either dimension and the logical structure of an inconsistency can be examined separately from its clinical content. The \emph{attribute} then names the specific clinical concept in conflict---a particular drug's dosing, a particular procedure's laterality---giving a fine-grained characterization within the domain.

\begin{table}[!htbp]
\centering
\caption{Taxonomy of clinical domains used to classify detected documentation inconsistencies.}
\label{tab:domain_taxonomy}
\footnotesize
\setlength{\tabcolsep}{4pt}
\renewcommand{\arraystretch}{1.2}
\begin{tabularx}{\linewidth}{l X}
\toprule
\textbf{Domain} & \textbf{Description} \\
\midrule
Demographics & Inconsistencies in patient-level attributes such as age, sex, or language, typically arising from copy-paste errors. \\
\addlinespace
Allergy & Inconsistencies between allergy lists and documented medication use, or conflicting allergy/sensitivity records. \\
\addlinespace
Procedure \& device & Contradictory claims about procedures performed (type, timing, findings) or the presence/status of implanted or indwelling devices. \\
\addlinespace
Diagnosis \& clinical finding & Conflicting diagnostic labels (e.g., HFpEF vs.\ HFrEF) or discrepant clinical observations (e.g., exam findings, vital signs, ejection fraction values). \\
\addlinespace
Lab \& Imaging & Discrepancies in reported laboratory values or imaging interpretations across different note sections. \\
\addlinespace
Medication & Contradictions involving drug names, dosages, routes, frequencies, or administration schedules across note sections. \\
\addlinespace
Disposition \& care planning & Conflicting documentation of functional status, discharge destination, follow-up plans, or goals-of-care directives. \\
\bottomrule
\end{tabularx}
\end{table}

\begin{table}[!htbp]
\centering
\caption{Taxonomy of logical contradiction axes used to classify the structural form of detected inconsistencies.}
\label{tab:axis_taxonomy}
\footnotesize
\setlength{\tabcolsep}{4pt}
\renewcommand{\arraystretch}{1.2}
\begin{tabularx}{\linewidth}{l X}
\toprule
\textbf{Contradiction axis} & \textbf{Description} \\
\midrule
Presence--absence & One section asserts that a clinical attribute exists while another denies or omits it (e.g., allergy documented vs.\ no known allergies). \\
\addlinespace
Status & Disagreement about the state of a clinical entity (e.g., active vs.\ resolved, current vs.\ discontinued, in situ vs.\ removed). \\
\addlinespace
Numeric value & Conflicting quantitative claims, such as differing laboratory values, dosage amounts, or measurements reported across sections. \\
\addlinespace
Timing & Inconsistent temporal assertions about when clinical events occurred (e.g., conflicting procedure dates, onset timing, or sequence of interventions). \\
\addlinespace
Laterality & Contradictory assertions about anatomical side (e.g., left vs.\ right kidney, ipsilateral vs.\ contralateral), with direct implications for procedural safety. \\
\addlinespace
Other & Less frequent structural forms, including identity mismatches, contraindication conflicts, and destination disagreements. \\
\bottomrule
\end{tabularx}
\end{table}

The present study should be interpreted as methodological and formative rather than as a definitive validation study. Its contribution is to expose the structure of the problem, identify recurring candidate inconsistencies and failure modes, and propose a testable ontology-schema framework. Definitive claims about sensitivity, precision, prevalence, or comparative model performance require the clinician-labeled corpus described in Section~\ref{sec:futurework}.

\subsection[Inconsistencies and Errors]{Inconsistencies and Errors}
\label{sec:errors}
The three classes relate to error in different ways. A strict contradiction entails a documentation error: the two statements cannot both be true, so at least one is wrong, although the note alone cannot say which. An ambiguity is a defect of clarity rather than of truth and need not involve an error at all. A gap is an error of omission: what is documented may be correct, but something the documentation requires is missing. Determining which statement of a contradiction is wrong, and whether an apparent omission reflects care that was given but not recorded, requires an independent account of the same admission, which the structured record can provide (Section~\ref{sec:crossvalidation}). Whether a documentation error then becomes a clinical error is a further question. Prior work links documentation discrepancies to downstream harm, particularly medication discrepancies at transitions of care and information propagated by copy-paste,\cite{cornish2005unintended,koppel2005cpoe,colicchio2019unintended} and several candidates in Table~\ref{tab:valid_examples} carry plausible safety consequences, such as a prescription for a drug listed as an allergy. We did not, however, measure whether any candidate reached a patient; linking candidates to outcomes would require clinical outcome data and is beyond the scope of this formative study.

\section[Limitations]{Limitations}
\label{sec:limitations}
This study has several limitations. First, all data come from MIMIC-IV, which reflects a single tertiary academic medical center, a patient population weighted toward intensive care, and documentation written before the spread of ambient AI scribes and other modern documentation tools; the kinds and frequency of candidates may differ at other institutions and in current documentation. Second, the 3{,}000 notes were drawn as three contiguous batches rather than at random, and because the corpus is ordered by patient, the batches contain repeated admissions of the same patient, so candidates are not independent. Third, the pipeline does not search exhaustively: it can miss inconsistencies, and without a reference standard we could not estimate its recall. Fourth, the internal chart review served a formative purpose: to characterize clinically meaningful inconsistencies and failure modes. Rigorous quantitative validation of the model’s ability to identify valid inconsistencies will require review by external clinicians. Fifth, detection used the discharge summary alone, although the same admission is also recorded in structured EHR data that could corroborate or refute a flagged statement. Finally, the pipeline used a single model family (Gemini 2.5), and all documentation was in English.

\section{Future Work}
\label{sec:futurework}

\subsection{A Clinician-Labeled Evaluation Corpus}
In this work, we developed a graded ontology and schema through empirical examination of pipeline output. The next step is to use this framework to construct a curated evaluation corpus, with full discharge summaries independently reviewed by clinical experts. Including summaries annotated without LLM assistance would support assessment of both how often the pipeline misses inconsistencies and how frequently its retained candidates are valid. Extending this evaluation to other health systems, whose documentation practices and EHR platforms differ from MIMIC-IV's, would test whether the findings generalize.

\subsection{Matching Methods to Failure Modes}
A second direction is to match refinement methods to the demands of each pipeline stage and the observed failure modes. Open-ended candidate identification requires broad exploration of the document, whereas verification requires focused reasoning about specific statements and their context. Future work should examine how LLMs can be combined with symbolic, graph-based, or purpose-built temporal methods. For example, explicit clinical-event timelines could help distinguish legitimate changes over time from documentation contradictions.

\subsection[Corroboration with Structured EHR Data]{Corroboration with Structured EHR Data}
\label{sec:crossvalidation}
Finally, the admission described in a discharge summary is also represented in structured EHR data, including laboratory results, medication orders, microbiology, diagnoses, and procedures. Comparing flagged statements with these records could help distinguish unclear documentation from statements that disagree with recorded care. Such corroboration, together with clinical adjudication, would support the next step toward linking documentation inconsistencies to documentation errors.
\section{Conclusion}
In this work, we conduct a formative study of the capability of LLMs to detect documentation inconsistencies in real-world EHRs, with a focus on discharge summaries, by developing a multi-stage LLM-based pipeline. Applied to MIMIC-IV, the pipeline surfaces candidate inconsistencies across medication, diagnosis, laboratory and imaging, allergy, procedural, and demographic content, many with potential implications for clinical reasoning, medication safety, and procedural care. At the same time, case-level inspection reveals recurring failure modes that motivate continued research on leveraging LLMs for inconsistency detection, discussed in detail in Sections~\ref{sec:lessons} and~\ref{sec:futurework}. Together, these findings establish a methodological foundation for automated detection of EHR documentation inconsistencies and a research agenda for moving from formative characterization toward rigorous, deployment-ready quality assurance for clinical documentation at scale.

\section*{Acknowledgments}
\noindent\textit{Use of AI tools.} Large language models (Google Gemini 2.5 Pro and Gemini 2.5 Flash) were used as analytical components of the inconsistency-detection pipeline described in the Methods. These models were accessed exclusively through Google Cloud Vertex AI batch prediction, a service PhysioNet lists as acceptable for use with credentialed data; under the Vertex AI terms, customer prompts and responses are not used to train Google's models. Separately, AI-assisted tools were used for limited proofreading and language polishing of the manuscript; all substantive content, analyses, interpretation, and final wording were reviewed and approved by the authors, who take full responsibility for the work.
\section*{Competing interests}

All authors declare no financial or non-financial competing interests.

\section*{Funding}

The research of J.L., M.T., W.W.S., and A.R.Z. was supported in part by NIH grant R01HL169347. The research of A.R.Z. was also supported in part by NIH grant R01HL168940.

\section*{Data Availability Statement}
The discharge summaries analyzed in this study were drawn from the
MIMIC-IV-Note module (v2.2) of the Medical Information Mart for Intensive
Care IV (MIMIC-IV) database, a publicly available, de-identified clinical
dataset distributed through PhysioNet under credentialed access. Access is
available to qualified researchers who complete the PhysioNet Credentialed
Health Data Use Agreement and the required CITI Data or Specimens Only
Research training. The dataset is referenced as:
\cite{johnson2020mimic,johnson2023mimicnote}. 

\section*{Contributorship Statement}
J.L. and A.R.Z. conceived the study. J.L., P.C., W.W.S., and A.R.Z. developed the methodology. J.L. and P.C. implemented the analysis pipeline and performed data processing. J.L. conducted the formal analysis and prepared the initial manuscript draft. M.T., R.B., C.W., W.W.S., D.Z., P.C., and A.R.Z. contributed to interpretation of the results and critical revision of the manuscript. M.T., R.B., C.W., W.W.S., and A.R.Z. provided clinical, informatics, and methodological expertise. J.L. prepared the figures and tables. M.T., W.W.S., and A.R.Z. supervised the study. All authors reviewed and approved the final manuscript.

\bibliography{sn-bibliography}

@inproceedings{gentimis2017predicting,
  title = {Predicting hospital length of stay using neural networks on mimic iii data},
  author = {Gentimis, T. and Alnaser, A. J. and Durante, Alex and Cook, Kyle and Steele, R.},
  booktitle = {2017 IEEE 15th intl conf on dependable, autonomic and secure computing, 15th intl conf on pervasive intelligence and computing, 3rd intl conf on big data intelligence and computing and cyber science and technology congress (DASC/PiCom/DataCom/CyberSciTech)},
  pages = {1194--1201},
  year = {2017},
  organization = {IEEE},
  journal = {2017 IEEE 15th Intl Conf on Dependable, Autonomic and Secure Computing, 15th Intl Conf on Pervasive Intelligence and Computing, 3rd Intl Conf on Big Data Intelligence and Computing and Cyber Science and Technology Congress(DASC/PiCom/DataCom/CyberSciTech)},
  doi = {10.1109/DASC-PICom-DataCom-CyberSciTec.2017.191},
  publisher = {IEEE},
}

@article{johnson2020mimic,
  title = {Mimic-iv},
  author = {Johnson, Alistair and Bulgarelli, Lucas and Pollard, Tom and Horng, Steven and Celi, Leo Anthony and Mark, Roger},
  journal = {PhysioNet. Available online at: https://physionet. org/content/mimiciv/1.0/(accessed August 23, 2021)},
  pages = {49--55},
  year = {2020},
}

@inproceedings{benabacha2025medec,
  title = {MEDEC: A Benchmark for Medical Error Detection and Correction in Clinical Notes},
  author = {Abacha, Asma Ben and Yim, Wen-wai and Fu, Yujuan and Sun, Zhaoyi and Yetisgen, Meliha and Xia, Fei and Lin, Thomas},
  booktitle = {Annual Meeting of the Association for Computational Linguistics},
  year = {2024},
  pages = {22539--22550},
  journal = {Annual Meeting of the Association for Computational Linguistics},
  doi = {10.48550/arXiv.2412.19260},
}

@inproceedings{wang2024mediqacorr,
  title = {WangLab at MEDIQA-CORR 2024: Retrieval-Augmented and DSPy-Optimized LLM Programs for Medical Error Detection and Correction},
  author = {Toma, Augustin and Xie, Ronald and Palayew, Steven and Lawler, Patrick and Wang, Bo},
  booktitle = {Proceedings of the 6th Clinical Natural Language Processing Workshop},
  year = {2024},
  address = {Mexico City, Mexico},
  publisher = {Association for Computational Linguistics},
  url = {https://aclanthology.org/2024.clinicalnlp-1.36},
  pages = {616-623},
  journal = {Proceedings of the 6th Clinical Natural Language Processing Workshop},
  doi = {10.18653/v1/2024.clinicalnlp-1.59},
}

@inproceedings{abacha2024mediqacorr,
  title = {Overview of the MEDIQA-CORR 2024 Shared Task on Medical Error Detection and Correction},
  author = {Abacha, Asma Ben and Yim, Wen-wai and Fu, Yujuan and Sun, Zhaoyi and Xia, Fei and Yetisgen, Meliha},
  booktitle = {Clinical Natural Language Processing Workshop},
  year = {2024},
  address = {Mexico City, Mexico},
  publisher = {Association for Computational Linguistics},
  pages = {596-603},
  url = {https://aclanthology.org/2024.clinicalnlp-1.34},
  journal = {Clinical Natural Language Processing Workshop},
  doi = {10.18653/v1/2024.clinicalnlp-1.57},
}

@inproceedings{gundabathula2024promptmind,
  title = {PromptMind at MEDIQA-CORR 2024: Improving Clinical Text Correction with Error Categorization and LLM Ensembles},
  author = {Gundabathula, Satya Kesav and Kolar, Sriram R.},
  booktitle = {Clinical Natural Language Processing Workshop},
  year = {2024},
  address = {Mexico City, Mexico},
  publisher = {Association for Computational Linguistics},
  journal = {Clinical Natural Language Processing Workshop},
  pages = {367-373},
  doi = {10.48550/arXiv.2405.08373},
}

@article{gema2024edinburgh,
  title = {Edinburgh clinical nlp at mediqa-corr 2024: Guiding large language models with hints},
  author = {Gema, Aryo Pradipta and Lee, Chaeeun and Minervini, Pasquale and Daines, Luke and Simpson, T Ian and Alex, Beatrice},
  journal = {Clinical Natural Language Processing Workshop},
  year = {2024},
  pages = {488-501},
  doi = {10.48550/arXiv.2405.18028},
  publisher = {Association for Computational Linguistics},
}

@article{yang2024enhancing,
  title = {Enhancing phenotype recognition in clinical notes using large language models: PhenoBCBERT and PhenoGPT},
  author = {Yang, Jingye and Liu, Cong and Deng, Wendy and Wu, Da and Weng, Chunhua and Zhou, Yunyun and Wang, Kai},
  journal = {Clinical Natural Language Processing Workshop},
  volume = {5},
  number = {1},
  year = {2023},
  publisher = {Elsevier},
  doi = {10.48550/arXiv.2308.06294},
}

@article{guo2024detection,
  title = {Detection of medication mentions and medication change events in clinical notes using transformer-based models},
  author = {Guo, Yuting and Ge, Yao and Sarker, Abeed},
  journal = {Medinfo},
  volume = {310},
  pages = {685},
  year = {2024},
  doi = {10.3233/SHTI231052},
  publisher = {IOS Press},
}

@article{harutyunyan2019multitask,
  title = {Multitask learning and benchmarking with clinical time series data},
  author = {Harutyunyan, Hrayr and Khachatrian, Hrant and Kale, David C. and Galstyan, A.},
  journal = {Scientific data},
  volume = {6},
  number = {1},
  pages = {96},
  year = {2017},
  publisher = {Nature Publishing Group UK London},
  doi = {10.1038/s41597-019-0103-9},
}

@article{komorowski2018artificial,
  title = {The artificial intelligence clinician learns optimal treatment strategies for sepsis in intensive care},
  author = {Komorowski, Matthieu and Celi, Leo A and Badawi, Omar and Gordon, Anthony C and Faisal, A Aldo},
  journal = {Nature medicine},
  volume = {24},
  number = {11},
  pages = {1716--1720},
  year = {2018},
  publisher = {Springer Science and Business Media LLC},
  doi = {10.1038/s41591-018-0213-5},
}

@article{bell2020patient,
  title = {Patient-reported errors in electronic health record ambulatory visit notes},
  author = {Bell, Susannah K. and Mejilla, Ralitza and Anselmo, Maria and Darer, Jonathan D. and Elmore, Joann G. and Leveille, Suzanne G. and Ngo, Long H. and Ralston, James D. and Delbanco, Tom and Walker, Jan},
  journal = {JAMA Network Open},
  volume = {3},
  number = {2},
  pages = {e200275},
  year = {2020},
  doi = {10.1001/jamanetworkopen.2020.0275},
  publisher = {American Medical Association (AMA)},
}

@article{weiskopf2013methods,
  title = {Methods and dimensions of electronic health record data quality assessment: enabling reuse for clinical research},
  author = {Weiskopf, Nicole G. and Weng, Chunhua},
  journal = {J. Am. Medical Informatics Assoc.},
  volume = {20},
  number = {1},
  pages = {144--151},
  year = {2013},
  doi = {10.1136/amiajnl-2011-000681},
  publisher = {Oxford University Press (OUP)},
}

@article{cornish2005unintended,
  title = {Unintended medication discrepancies at the time of hospital admission},
  author = {Cornish, P. and Knowles, S. and Marchesano, Romina and Tam, Vincent and Shadowitz, S. and Juurlink, D. and Etchells, E.},
  journal = {Archives of Internal Medicine},
  volume = {165},
  number = {4},
  pages = {424},
  year = {2005},
  doi = {10.1001/archinte.165.4.424},
  publisher = {American Medical Association (AMA)},
}

@misc{jointcommission2018copy,
  title = {Safe use of health information technology},
  author = {{The Joint Commission}},
  howpublished = {Sentinel Event Alert 58},
  year = {2015},
  url = {https://digitalassets.jointcommission.org/api/public/content/2dc4cad7bdc34451b85a2be5431f09ef?v=c9afdcd7},
  journal = {Sentinel event alert},
}

@misc{jointcommission2014copypaste,
  title = {Safe Practices for Copy and Paste in the Electronic Health Record},
  author = {{The Joint Commission}},
  howpublished = {Sentinel Event Alert, Issue 54},
  year = {2014},
  journal = {Applied Clinical Informatics},
  volume = {26},
  number = {01},
  pages = {12-34},
  doi = {10.4338/aci-2016-09-r-0150},
  publisher = {Georg Thieme Verlag KG},
}

@article{Hogan2015,
  title = {Preventable deaths due to problems in care in English acute hospitals: a retrospective case record review study},
  author = {Hogan, H. and Healey, F. and Neale, G. and Thomson, R. and Vincent, C. and Black, N.},
  journal = {BMJ Quality \& Safety},
  volume = {21},
  year = {2012},
  number = {9},
  pages = {737-745},
  doi = {10.1136/bmjqs-2011-001159},
  publisher = {BMJ},
}

@book{balogh2015improving,
  editor = {Balogh, Erin P. and Miller, Bryan T. and Ball, John R.},
  title = {Improving Diagnosis in Health Care},
  publisher = {National Academies Press},
  address = {Washington, DC},
  year = {2015},
  doi = {10.17226/21794},
  isbn = {978-0-309-37769-0},
  note = {Committee on Diagnostic Error in Health Care, Board on Health Care Services, Institute of Medicine, National Academies of Sciences, Engineering, and Medicine},
}

@article{graber2017impact,
  title = {The impact of electronic health records on diagnosis},
  author = {Graber, Mark L. and Byrne, Colleen and Johnston, Douglas},
  journal = {Diagnosis},
  volume = {4},
  number = {4},
  pages = {211--223},
  year = {2017},
  doi = {10.1515/dx-2017-0012},
  publisher = {Walter de Gruyter GmbH},
}

@article{koppel2005cpoe,
  author = {Koppel, Ross and Metlay, Joshua P. and Cohen, Abigail and Abaluck, Brian and Localio, A. Russell and Kimmel, Stephen E. and Strom, Brian L.},
  title = {Role of Computerized Physician Order Entry Systems in Facilitating Medication Errors},
  journal = {JAMA},
  volume = {293},
  number = {10},
  pages = {1197--1203},
  year = {2005},
  doi = {10.1001/jama.293.10.1197},
}

@article{colicchio2019unintended,
  title = {Unintended Consequences of Nationwide Electronic Health Record Adoption: Challenges and Opportunities in the Transition to a Digital Health System},
  author = {Colicchio, Tiago K. and Cimino, James J. and Del Fiol, Guilherme},
  journal = {Journal of Medical Internet Research},
  volume = {21},
  number = {6},
  pages = {e13313},
  year = {2019},
  doi = {10.2196/13313},
  publisher = {JMIR Publications Inc.},
}

@article{weir2003direct,
  title = {Direct Text Entry in Electronic Progress Notes: An Evaluation of Input Errors},
  author = {Weir, Charlene R. and Hurdle, John F. and Felgar, Marjorie A. and Hoffman, Jeanette M. and Roth, Bruce and Nebeker, Jonathan R.},
  journal = {Methods of Information in Medicine},
  volume = {42},
  number = {1},
  pages = {61--67},
  year = {2003},
  doi = {10.1055/s-0038-1634210},
}

@article{singh2014frequency,
  title = {The frequency of diagnostic errors in outpatient care: estimations from three large observational studies involving US adult populations},
  author = {Singh, Hardeep and Meyer, Ashley N. D. and Thomas, Eric J.},
  journal = {BMJ Quality \& Safety},
  volume = {23},
  number = {9},
  pages = {727--731},
  year = {2014},
  doi = {10.1136/bmjqs-2013-002627},
  publisher = {BMJ},
}

@article{google2025gemini25,
  title = {Gemini 2.5: Pushing the Frontier with Advanced Reasoning, Multimodality, Long Context, and Next Generation Agentic Capabilities},
  author = {Comanici, Gheorghe and Bieber, E. and Schaekermann, Mike and Pasupat, Ice and Sachdeva, Noveen and Dhillon, Inderjit S. and Blistein, Marcel and Ram, Ori and Zhang, Dan and Rosen, Evan and others},
  journal = {arXiv.org},
  year = {2025},
  url = {https://arxiv.org/abs/2507.06261},
}

@article{guha2023legalbench,
  title = {{LegalBench}: A Collaboratively Built Benchmark for Measuring Legal Reasoning in Large Language Models},
  author = {Guha, Neel and Nyarko, Julian and Ho, Daniel E. and Ré, Christopher and Chilton, Adam and Narayana, Aditya and Chohlas-Wood, Alex and Peters, Austin M. K. and Waldon, Brandon and Rockmore, D. and others},
  journal = {arXiv preprint},
  year = {2023},
  note = {arXiv:2308.11462},
  url = {https://arxiv.org/abs/2308.11462},
}

@article{wu2023bloomberggpt,
  title = {{BloombergGPT}: A Large Language Model for Finance},
  author = {Wu, Shijie and Irsoy, Ozan and Lu, Steven and Dabravolski, Vadim and Dredze, Mark and Gehrmann, Sebastian and Kambadur, Prabhanjan and Rosenberg, David and Mann, Gideon},
  journal = {arXiv preprint},
  year = {2023},
  note = {arXiv:2303.17564},
  url = {https://arxiv.org/abs/2303.17564},
}

@article{thirunavukarasu2023llmmedicine,
  title = {Large language models in medicine},
  author = {Thirunavukarasu, Arun James and Ting, Darren Shu Jeng and Elangovan, Kabilan and Gutierrez, Laura and Tan, Ting Fang and Ting, Daniel Shu Wei},
  journal = {Nature Medicine},
  volume = {29},
  number = {8},
  pages = {1930--1940},
  year = {2023},
  publisher = {Springer Science and Business Media LLC},
  doi = {10.1038/s41591-023-02448-8},
}

@article{clusmann2023future,
  title = {The future landscape of large language models in medicine},
  author = {Clusmann, J. and Kolbinger, F. and Muti, H. and Carrero, Zunamys I. and Eckardt, Jan-Niklas and Laleh, Narmin Ghaffari and Löffler, C. and Schwarzkopf, Sophie-Caroline and Unger, Michaela and Veldhuizen, G. P. and others},
  journal = {Communications Medicine},
  volume = {3},
  number = {1},
  pages = {141},
  year = {2023},
  publisher = {Springer Science and Business Media LLC},
  doi = {10.1038/s43856-023-00370-1},
}

@article{singhal2023medpalm,
  title = {Large language models encode clinical knowledge},
  author = {Singhal, K. and Azizi, Shekoofeh and Tu, T. and Mahdavi, S. and Wei, Jason and Chung, Hyung Won and Scales, Nathan and Tanwani, A. and Cole-Lewis, H. and Pfohl, S. and others},
  journal = {Nature},
  volume = {620},
  number = {7972},
  pages = {172--180},
  year = {2022},
  publisher = {Nature Publishing Group},
  doi = {10.1038/s41586-023-06291-2},
}

@article{singhal2025medpalm2,
  title = {Toward expert-level medical question answering with large language models},
  author = {Singhal, Karan and Tu, Tao and Gottweis, Juraj and Sayres, R. and Wulczyn, Ellery and Amin, Mohamed and Hou, Le and Clark, Kevin and Pfohl, Stephen R. and Cole-Lewis, Heather and others},
  journal = {Nature Medicine},
  year = {2025},
  publisher = {Springer Science and Business Media LLC},
  url = {https://arxiv.org/abs/2305.09617},
  volume = {31},
  number = {3},
  pages = {943-950},
  doi = {10.1038/s41591-024-03423-7},
}

@inproceedings{agrawal2022clinicalIE,
  title = {Large Language Models are Few-Shot Clinical Information Extractors},
  author = {Agrawal, Monica and Hegselmann, Stefan and Lang, Hunter and Kim, Yoon and Sontag, David},
  booktitle = {Conference on Empirical Methods in Natural Language Processing},
  year = {2022},
  url = {https://aclanthology.org/2022.emnlp-main.130/},
  journal = {Conference on Empirical Methods in Natural Language Processing},
  pages = {1998-2022},
  doi = {10.18653/v1/2022.emnlp-main.130},
  publisher = {Association for Computational Linguistics},
}

@article{yang2022gatortron,
  title = {A large language model for electronic health records},
  author = {Yang, Xi and Chen, Aokun and Pournejatian, Nima M. and Shin, Hoo-Chang and Smith, Kaleb E. and Parisien, Christopher and Compas, Colin B. and Martin, Cheryl and Costa, Anthony B and Flores, Mona G. and others},
  journal = {npj Digital Medicine},
  volume = {5},
  number = {1},
  pages = {194},
  year = {2022},
  publisher = {Springer Science and Business Media LLC},
  doi = {10.1038/s41746-022-00742-2},
}

@article{vanveen2024clinicalsumm,
  title = {Adapted large language models can outperform medical experts in clinical text summarization},
  author = {Veen, Dave Van and Uden, Cara Van and Blankemeier, L. and Delbrouck, Jean-Benoit and Aali, Asad and Blüthgen, Christian and Pareek, A. and Polacin, Malgorzata and Collins, William and Ahuja, Neera and others},
  journal = {Nature Medicine},
  volume = {30},
  number = {4},
  pages = {1134--1142},
  year = {2023},
  publisher = {Nature Publishing Group},
  doi = {10.1038/s41591-024-02855-5},
}

@inproceedings{wei2022cot,
  title = {Chain-of-Thought Prompting Elicits Reasoning in Large Language Models},
  author = {Wei, Jason and Wang, Xuezhi and Schuurmans, Dale and Bosma, Maarten and Chi, Ed H. and Xia, F. and Le, Quoc and Zhou, Denny},
  booktitle = {Neural Information Processing Systems (NeurIPS)},
  year = {2022},
  pages = {24824-24837},
}

@book{dowden,
  author    = {Dowden, Bradley H.},
  title     = {Logical Reasoning},
  publisher = {California State University, Sacramento},
  year      = {2015},
  note      = {Open-access textbook},
}

@article{johnson2023mimicnote,
  author = {Johnson, Alistair and Pollard, Tom and Horng, Steven and Celi, Leo Anthony and Mark, Roger},
  title = {MIMIC-IV-Note: Deidentified Free-Text Clinical Notes},
  journal = {PhysioNet},
  year = {2023},
  note = {Version 2.2},
  doi = {10.13026/1n74-ne17}
}
\end{document}